\documentclass[letterpaper]{article} 
\usepackage{aaai2026}  
\usepackage{times}  
\usepackage{helvet}  
\usepackage{courier}  
\usepackage[hyphens]{url}  
\usepackage{graphicx} 
\usepackage{natbib}  
\usepackage{caption} 
\usepackage{algorithm}
\usepackage{algorithmic}
\usepackage{enumitem}

\usepackage{newfloat}
\usepackage{listings}
\DeclareCaptionStyle{ruled}{labelfont=normalfont,labelsep=colon,strut=off} 
\floatstyle{ruled}
\newfloat{listing}{tb}{lst}{}
\floatname{listing}{Listing}
\title{Federated Learning Playground}

\author {
    Bryan Guanrong Shan\textsuperscript{\rm 1},
    Alysa Ziying Tan\textsuperscript{\rm 1,2,3},
    Han Yu\textsuperscript{\rm 1}
}
\affiliations {
    \textsuperscript{\rm 1}Nanyang Technological University (NTU), Singapore\\
    \textsuperscript{\rm 2}Alibaba-NTU Singapore Joint Research Institute, NTU, Singapore \\ \textsuperscript{\rm 3}Alibaba Group, Hangzhou, China \\
    \{bshan001,s190109,han.yu\}@ntu.edu.sg
}
\begin{document}
\maketitle
\begin{abstract}
We present \textit{Federated Learning Playground}, an interactive browser-based platform inspired by and extends TensorFlow Playground that teaches core Federated Learning (FL) concepts. Users can experiment with heterogeneous client data distributions, model hyperparameters, and aggregation algorithms directly in the browser without coding or system setup, and observe their effects on client and global models through real-time visualizations, gaining intuition for challenges such as non-IID data, local overfitting, and scalability. The playground serves as an easy to use educational tool, lowering the entry barrier for newcomers to distributed AI while also offering a sandbox for rapidly prototyping and comparing FL methods. By democratizing exploration of FL, it promotes broader understanding and adoption of this important paradigm.
\end{abstract}

\section{Introduction}
Federated learning (FL) is a collaborative machine learning paradigm that enables multiple clients to train a shared model without exchanging raw data. This paradigm enables collaboration without exposing raw data, making it especially valuable in domains such as healthcare, finance, and mobile intelligence where privacy is critical (\citealt{kairouz2021advances,hard2019federatedlearningmobilekeyboard}). 

Despite its growing importance, FL introduces unique challenges that distinguish it from traditional centralized learning. These include handling non-independent and identically distributed (non-IID) client data, partial client participation, and variability in systems and resources (\citealt{tan2023towards}). Understanding the effects of these challenges on different algorithmic and system-level choices often requires coding expertise and access to specialized infrastructure, which discourage newcomers from experimenting with FL.



In the broader machine learning community, open-source tools such as TensorFlow Playground (\citealt{smilkov2017directmanipulationvisualizationdeepnetworks}) have shown the power of interactive visualization for lowering entry barriers. By allowing users to tinker with small neural networks directly in the browser, it has helped learners and practitioners intuitively grasp the effects of hyperparameters and model choices without writing code. This combination of simplicity and interactivity has made it a widely used educational and demonstration resource.

Motivated by this success, we present the \textit{Federated Learning Playground}, an interactive, browser-based platform that brings the same spirit of accessibility to the domain of FL. Users can adjust data distributions, hyperparameters, and aggregation strategies, then observe the effects on client and global models through real-time visualizations. The playground serves as both an educational tool and a sandbox that lowers the entry barrier for newcomers exploring privacy-preserving and distributed AI.




\section{System Design and Implementation}
The system is built around an FL engine and orchestration layer that extends the standard centralized training loop in TensorFlow Playground. Implemented as \texttt{oneStepFL}, this step (1) samples a fraction of clients, (2) performs local training for a configurable number of epochs, and (3) performs server-side aggregation. When the ``Run Federated'' option is disabled, the system falls back to the standard centralized training mode, enabling learners to compare the two modes side by side. The entire implementation runs client-side, making it lightweight and easily deployable in a web browser without external dependencies. Combined with a clear and intuitive interface, this design ensures that the \textit{Federated Learning Playground} remains accessible, interactive, and extensible.


\textbf{FL Aggregation Algorithms.}
Users can select:
\begin{itemize}[leftmargin=2em]
\item \textbf{FedAvg} \citep{mcmahan2017communication}. Weighted averaging of client deltas by data size.
\item \textbf{FedProx} \citep{li2020federatedprox}. Adds a proximal term to stabilize local updates under data heterogeneity.
\item \textbf{FedAdam} \citep{reddi2021adaptive}. Applies Adam optimization at the server for adaptive aggregation.
\item \textbf{SCAFFOLD} \citep{karimireddy2020scaffold}. Uses control variates to correct client drift on non-IID data.
\end{itemize}

\textbf{FL Hyperparameters.} Users can tune the following:

\begin{itemize}[leftmargin=2em]
\item \textbf{Non-IID partitioning.}
Data are split across $N$ virtual clients using a Dirichlet-based sampler to control the degree of heterogeneity (see Figure 1).

\item\textbf{Clustered FL.}
Supports $k$-means clustering (cosine or $\ell_2$) of client updates, illustrating the benefits of grouping when client distributions are multi-modal.


\item\textbf{Differential privacy.}
Each client update is clipped to a norm bound and perturbed with Gaussian noise before aggregation, following DP-SGD (\citealt{abadi2016deep}). Visualizations illustrate the privacy–utility trade-off as the noise level varies.

\item\textbf{Client behavior.}
Users can adjust probability of dropout, client learning rate and epochs. Updates are real-time to illustrate impact on communication costs.

\item\textbf{Local model training.}
Training of individual clients can be run for specific clients to visualize the effect of local training and global model aggregation.
\end{itemize}

\begin{figure}[t]
  \centering
  \includegraphics[width=\linewidth]{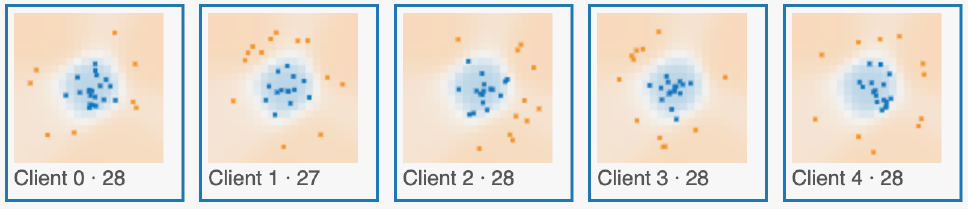}\par\vspace{0em}
  \includegraphics[width=\linewidth]{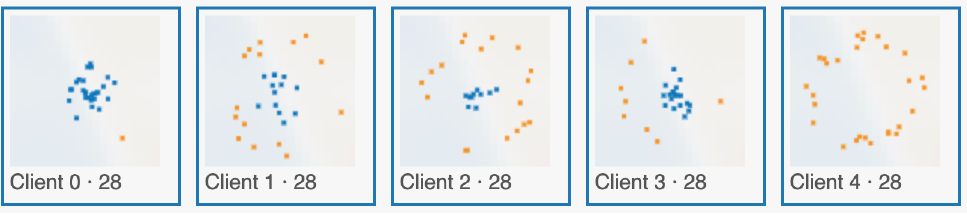}\par\vspace{0em}
  \includegraphics[width=\linewidth]{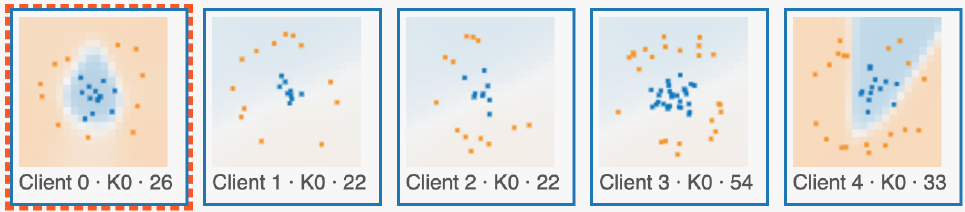}
  \caption{Client Data Distributions. \textbf{Top}: IID distribution across clients (post-training).
  \textbf{Middle}: Non-IID distribution (pre-training).
  \textbf{Bottom}: Uniform class distribution with skewed sample sizes, after local training of clients 0 and 4.}
  \label{fig:client_data_dist}
\end{figure}

Integrating FL support has expanded the codebase to more than twice the size of the original TensorFlow Playground, with the increase coming from additional modules for client sampling, local training, server aggregation, FL algorithms, and privacy mechanisms. Key extensions include:
\paragraph{Weight representation.} Weights and Bias are flattened into a single vector (see \texttt{nnFlattenWeights}), enabling efficient computation of deltas, proximal terms, control variates, and DP noise in a single pass.

\paragraph{Round orchestration.} \texttt{oneStepFL} samples clients, runs local epochs with optional FedProx or SCAFFOLD, forms deltas, applies DP clipping/noise, and aggregates. FedAdam is supported via a lightweight server-side Adam optimizer.

\paragraph{Clustering.} Clustered FL initializes $K$ models and performs $k$-means clustering every $n$ rounds after warmup, with client weights averaged within each cluster.

\section{User Interface and Visualizations}
The UI adds FL-specific controls, i.e. clients no. \& fraction, local epochs, non-IID ($\alpha$), clustering ($k$), differential privacy with clipping/noise, and algorithm selector (Figure 2). 

We provide visualizations (Figure 3) for:
\begin{enumerate}
  \item \textbf{Client participation} for sampling fairness \& coverage.
  \item \textbf{Communication cost} for efficiency comparisons.
  \item \textbf{Client loss distribution} for heterogeneity \& drift.
  \item \textbf{Convergence Rate} for training progress.
  \item \textbf{Client Data distribution} for data spread.
\end{enumerate}

\section{Reproducibility and Artifact}
The system is released as a static web application hosted on GitHub Pages at \url{https://oseltamivir.github.io/playground}. The code is open-sourced, with documentation.

\begin{figure}[t]
  \centering
  \setlength{\fboxsep}{0pt}
  \fbox{\includegraphics[width=0.99\columnwidth]{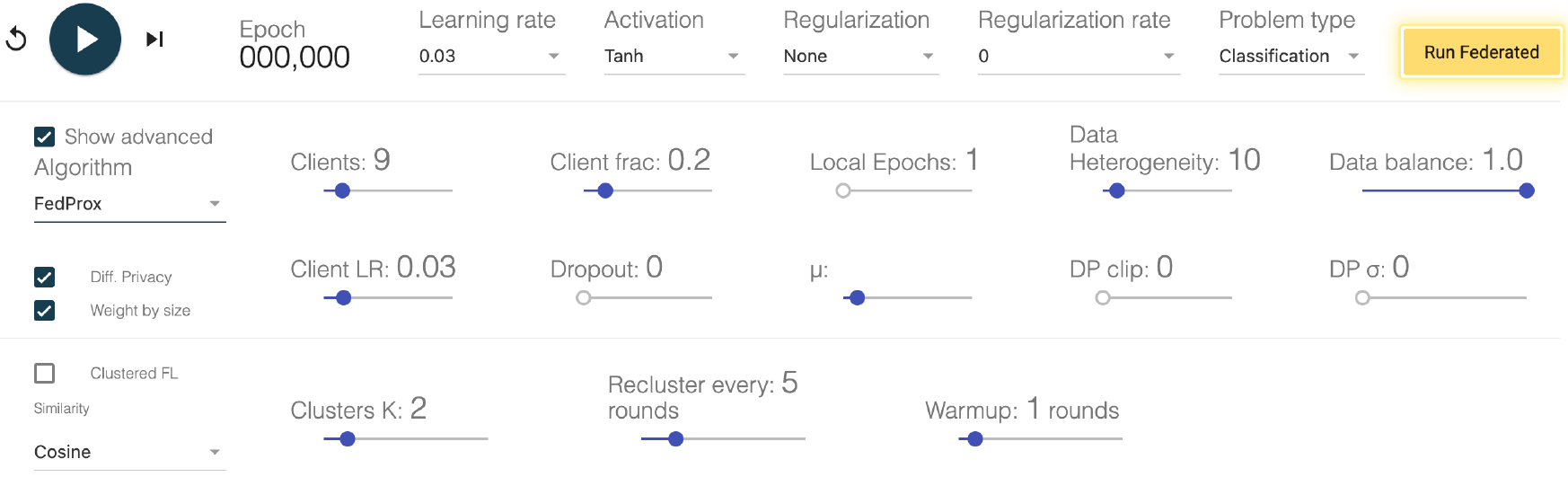}}
  \caption{Overview of interface controls. FL-specific controls in Rows 2–4, with original controls retained in Row 1.
  }
  \label{fig:overview}
\end{figure}

\begin{figure}[t]
  \centering
  \includegraphics[width=1\columnwidth]{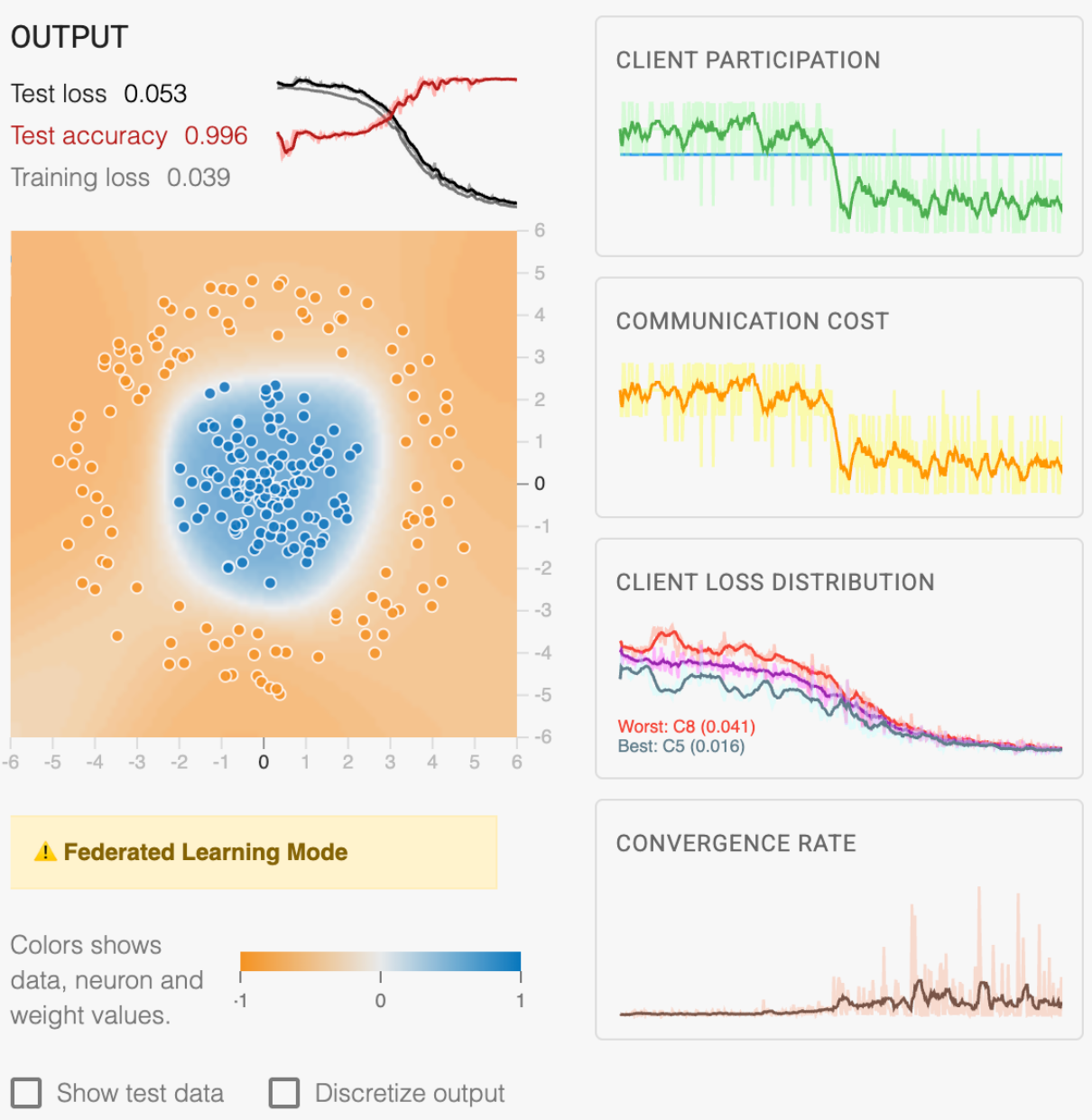}
  \caption{Simulated increase in client dropout halfway through training, with visualizations of \textbf{Client participation}, 
  \textbf{Comms cost} \textit{(scales with no. of clients)}, \textbf{Client loss distribution} \textit{(max, mean, min)}, and \textbf{Convergence rate}.}
  \label{fig:metrics}
\end{figure}


\section{Conclusions and Future Work}
The \textit{Federated Learning Playground} is a lightweight didactic tool using tiny MLPs and synthetic 2D data to illustrate core FL concepts accessibly. It bridges education and research in FL for students and researchers alike, lowering barriers and inspiring exploration of distributed-AI challenges. Future work will add trustworthy-FL features for explainability and fairness, and broaden the simulator to vertical FL and FL foundation models. (\citealt{ren2025advances}).

\section*{Acknowledgments}
This work is supported, in part, by the Alibaba Group through Alibaba Innovative Research (AIR) Program and Alibaba-NTU Singapore Joint Research Institute (JRI) (Alibaba-NTU-AIR2019B1); and Nanyang Technological University, Singapore; the Ministry of Education, Singapore, under its Academic Research Fund Tier 1 (RG101/24), and conducted as part of the IMDA SG Digital Leadership Accelerator Mentorship Program for Young Talent.

\bibliography{aaai2026}

@misc{hard2019federatedlearningmobilekeyboard,
      title={Federated Learning for Mobile Keyboard Prediction}, 
      author={Andrew Hard and Kanishka Rao and Rajiv Mathews and Swaroop Ramaswamy and Françoise Beaufays and Sean Augenstein and Hubert Eichner and Chloé Kiddon and Daniel Ramage},
      year={2019},
      eprint={1811.03604},
      archivePrefix={arXiv},
      primaryClass={cs.CL},
      url={https://arxiv.org/abs/1811.03604}, 
}

@misc{smilkov2017directmanipulationvisualizationdeepnetworks,
      title={Direct-Manipulation Visualization of Deep Networks}, 
      author={Daniel Smilkov and Shan Carter and D. Sculley and Fernanda B. Viégas and Martin Wattenberg},
      year={2017},
      eprint={1708.03788},
      archivePrefix={arXiv},
      primaryClass={cs.LG},
      url={https://arxiv.org/abs/1708.03788}, 
}

@inproceedings{mcmahan2017communication,
  title={Communication-Efficient Learning of Deep Networks from Decentralized Data},
  author={McMahan, H. Brendan and Moore, Eider and Ramage, Daniel and Hampson, Seth and Arcas, Blaise Aguera y},
  booktitle={AISTATS},
  year={2017}
}

@inproceedings{li2020federatedprox,
  title={Federated Optimization in Heterogeneous Networks},
  author={Li, Tian and Sahu, Anit Kumar and Talwalkar, Ameet and Smith, Virginia},
  booktitle={PMLR},
  year={2020}
}

@inproceedings{reddi2021adaptive,
  title={Adaptive Federated Optimization},
  author={Reddi, Sashank and Charles, Zachary and Zaheer, Manzil and others},
  booktitle={ICLR},
  year={2021}
}

@inproceedings{karimireddy2020scaffold,
  title={SCAFFOLD: Stochastic Controlled Averaging for Federated Learning},
  author={Karimireddy, Sai Praneeth and Kale, Satyen and Mohri, Mehryar and Reddi, Sashank and Stich, Sebastian and Suresh, Ananda Theertha},
  booktitle={ICML},
  year={2020}
}

@inproceedings{abadi2016deep,
  title={Deep Learning with Differential Privacy},
  author={Abadi, Martin and Chu, Andy and Goodfellow, Ian and others},
  booktitle={CCS},
  year={2016}
}

@article{tan2023towards,
  title={Towards Personalized Federated Learning},
  author={Tan, Alysa Ziying and Yu, Han and Cui, Lizhen and Yang, Qiang},
  journal={IEEE TNNLS},
  year={2023},
  publisher={IEEE},
  volume = {34},
  number = {12},
  pages = {9587–9603}
}

@article{kairouz2021advances,
  title={Advances and open problems in federated learning},
  author={Kairouz, Peter and McMahan, H Brendan and Avent, Brendan and Bellet, Aur{\'e}lien and Bennis, Mehdi and Bhagoji, Arjun Nitin and Bonawitz, Kallista and Charles, Zachary and Cormode, Graham and Cummings, Rachel and others},
  journal={Found. Trends Mach. Learn.},
  volume={14},
  number={1--2},
  pages={1--210},
  year={2021}
}

@article{ren2025advances,
  title={Advances and open challenges in federated learning with foundation models},
  author={Ren, Chao and Yu, Han and Peng, Hongyi and Tang, Xiaoli and Li, Anran and Gao, Yulan and Tan, Alysa Ziying and Zhao, Bo and Li, Xiaoxiao and Li, Zengxiang and others},
  journal={IEEE COMST},
  year={2025}
}
\end{document}